%% file: paper.tex
\documentclass[letterpaper]{article} 
\usepackage[preprint]{aaai2027}  
\usepackage[hyphens]{url}  
\usepackage{graphicx} 
\usepackage{natbib}  
\usepackage{caption} 
\usepackage{algorithm}
\usepackage{algorithmic}

\usepackage{amsmath}
\usepackage{amssymb}
\usepackage{arydshln}
\usepackage{multirow}
\usepackage{tabularx}
\usepackage{enumitem}
\usepackage{array}
\usepackage{makecell}
\usepackage{newfloat}
\usepackage{listings}
\DeclareCaptionStyle{ruled}{labelfont=normalfont,labelsep=colon,strut=off} 
\floatstyle{ruled}
\newfloat{listing}{tb}{lst}{}
\floatname{listing}{Listing}

\usepackage{booktabs}

\title{CanonNav: Disentangling Navigation Behavior from Camera Geometry in Cross-Platform Visual Navigation}
\author{
    Dong-Wook Kim, Ji-Hoon Hwang, E-In Son, Mintaek Oh, Seung-Woo Seo\corresponding
}
\affiliations{
    Seoul National University, Republic of Korea
}

\begin{document}

\maketitle

\begin{abstract}
While visual navigation has advanced through imitation learning from cross-platform demonstrations, fully leveraging such data remains challenging. First, directly learning from image-trajectory pairs entangles navigation behavior with platform-dependent camera geometry. This hinders consistent learning by forcing the policy to implicitly infer camera geometry from visual observations, an inherently ill-posed problem. Second, imitation learning from demonstrated trajectories captures the expert's chosen motion but leaves the intermediate decisions underlying that motion implicit. To address these issues, we propose CanonNav, a visual navigation framework that disentangles navigation behavior from camera geometry and incorporates complementary planning supervision into learning from cross-platform demonstrations. CanonNav introduces camera geometry canonicalization, which transforms visual observations and trajectories into a camera-consistent representation space. Building on this representation, we derive safety and local-progress supervision using pseudo-labels from an offline traversability estimator. Safety supervision penalizes unsafe trajectories, while local-progress supervision guides where the robot should advance. Experiments across diverse camera configurations and environments show that, despite using only RGB at inference, CanonNav consistently outperforms RGB-based baselines and even surpasses RGB-D-based methods in challenging scenarios.
\end{abstract}


\begin{figure}[t]
    \centering
    \includegraphics[width=\linewidth]{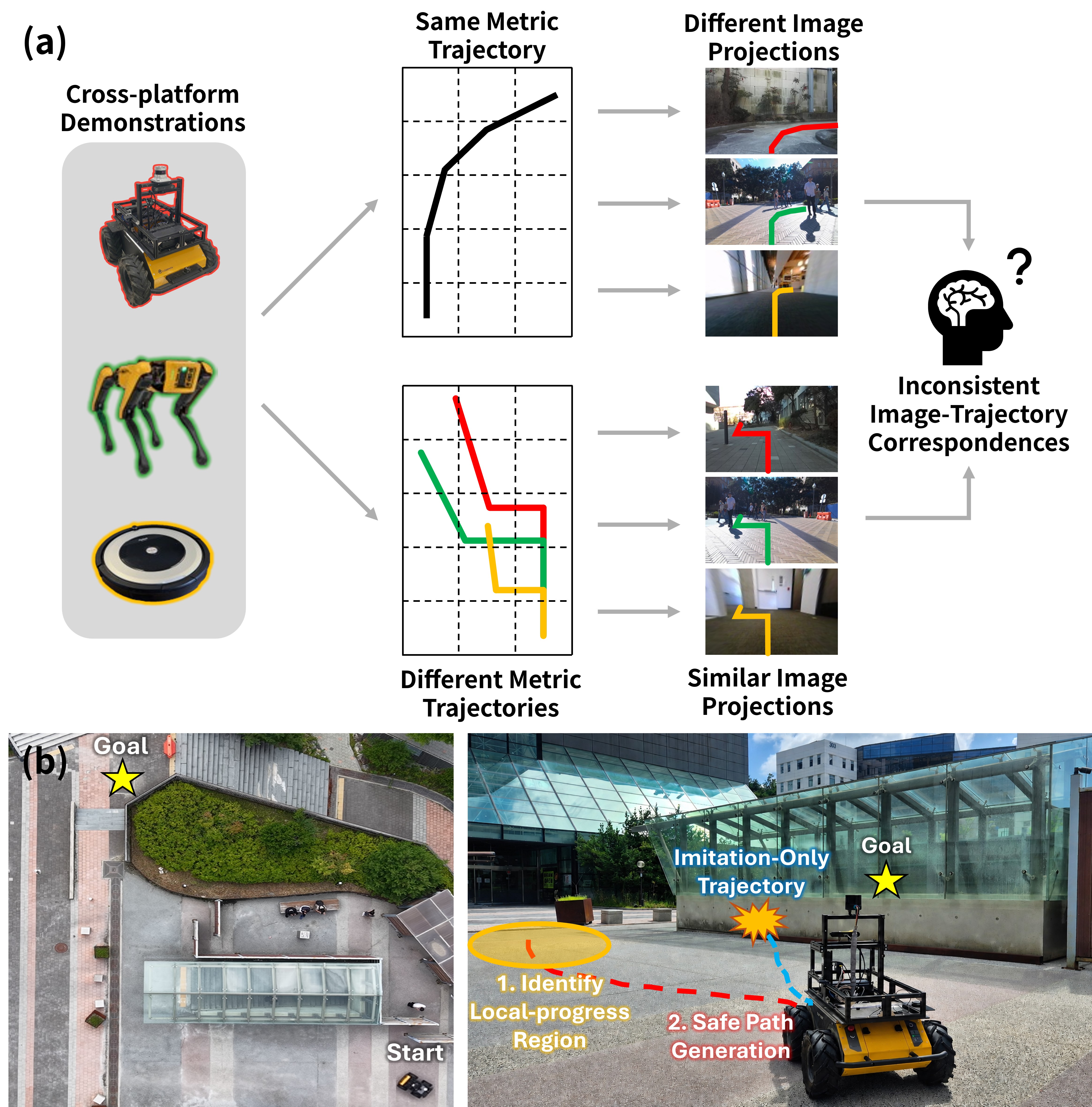}
    \caption{Motivation of CanonNav. (a) Different camera geometries induce inconsistent image-trajectory correspondences, hindering consistent policy learning across platforms. (b) Imitation learning provides limited supervision for intermediate planning decisions, such as identifying a local-progress region and generating a safe path toward it.}
    \label{fig:intro}
\end{figure}

\section{Introduction}
Visual navigation enables robots to reach a specified goal using onboard visual observations. Recent learning-based approaches~\citep{shah2023gnm, shah2023vint, sridhar2024nomad} have achieved notable performance by scaling imitation learning with large navigation datasets~\citep{shah2021rapid, karnan2022socially, hirose2023sacson, hirose2025learning} and simulation platforms~\citep{savva2019habitat, makoviychuk2021isaac}. As policies are trained on diverse cross-platform data, a central question emerges: What is needed to fully leverage these heterogeneous demonstrations for policy learning? We argue that this requires establishing a canonical representation space that disentangles navigation behavior from platform-dependent camera geometry and enables supervision signals to be applied consistently across such demonstrations.

Existing methods typically learn to map visual observations directly to robot-centric trajectories via imitation learning. However, this approach neglects that the correspondence between images and trajectory labels depends on camera geometry. As illustrated in Fig.~\ref{fig:intro}(a), the same robot-centric trajectory label is associated with different image regions across different platforms. The policy is therefore forced to implicitly infer camera geometry from each observation while learning navigation behavior. Since this inference is inherently ill-posed, coupling it with policy learning hinders consistent learning across platforms. Prior methods~\citep{hirose2023exaug, chen2024roviaug, eftekhar2024one} mitigate this issue by viewpoint augmentation or embodiment randomization. However, expanding the training distribution does not fundamentally resolve camera-dependent correspondence, limiting generalization to variations covered during training. Learning from cross-platform demonstrations therefore calls for a canonical representation space that aligns visual observations and trajectories across platforms.

While such a canonical representation mitigates geometric inconsistency in trajectory imitation, leveraging heterogeneous demonstrations further requires explicit planning supervision. As shown in Fig.~\ref{fig:intro}(b), when the goal cannot be reached directly, the policy must determine where to advance locally and how to reach that region safely. However, demonstrated trajectories reveal only the expert's chosen motion and leave these intermediate decisions implicit. Prior methods~\citep{roth2024viplanner, cai2025navdp} use depth inputs and training signals to provide such cues, but depth is unavailable in RGB-only deployment, and providing such cues consistently across camera configurations remains underexplored. This motivates designing planning supervision within the canonical representation so that it can complement trajectory imitation consistently across platforms.

To address these challenges, we propose CanonNav, a visual navigation framework that disentangles navigation behavior from camera geometry and incorporates complementary planning supervision into learning from cross-platform demonstrations. CanonNav introduces camera geometry canonicalization, which transforms observations and trajectories into a camera-consistent representation space. Building on this representation, we derive safety and local-progress supervision using pseudo-labels from an offline traversability estimator. Safety supervision penalizes unsafe trajectories, while local-progress supervision guides the prediction of the Scope-of-Reach, which we define as a local region where the robot should advance. Within the canonicalized space, these signals enable consistent learning of collision-aware and goal-directed navigation behavior from cross-platform demonstrations. Simulation and real-world experiments across camera configurations and environments show that, despite RGB-only inference, CanonNav consistently outperforms RGB-based baselines and even surpasses RGB-D-based methods in challenging scenarios.

Our contributions are summarized as follows:
\begin{itemize}
    \item We introduce camera geometry canonicalization to establish a camera-consistent representation space that disentangles navigation behavior from camera geometry.

    \item Within this space, we derive training-time safety and local-progress supervision from offline traversability estimation for collision-aware and goal-directed planning.

    \item Extensive experiments show that CanonNav consistently outperforms RGB baselines and surpasses RGB-D methods in challenging scenarios with RGB-only inference.
\end{itemize}

\begin{figure*}[t]
    \centering
    \includegraphics[width=0.9\textwidth]{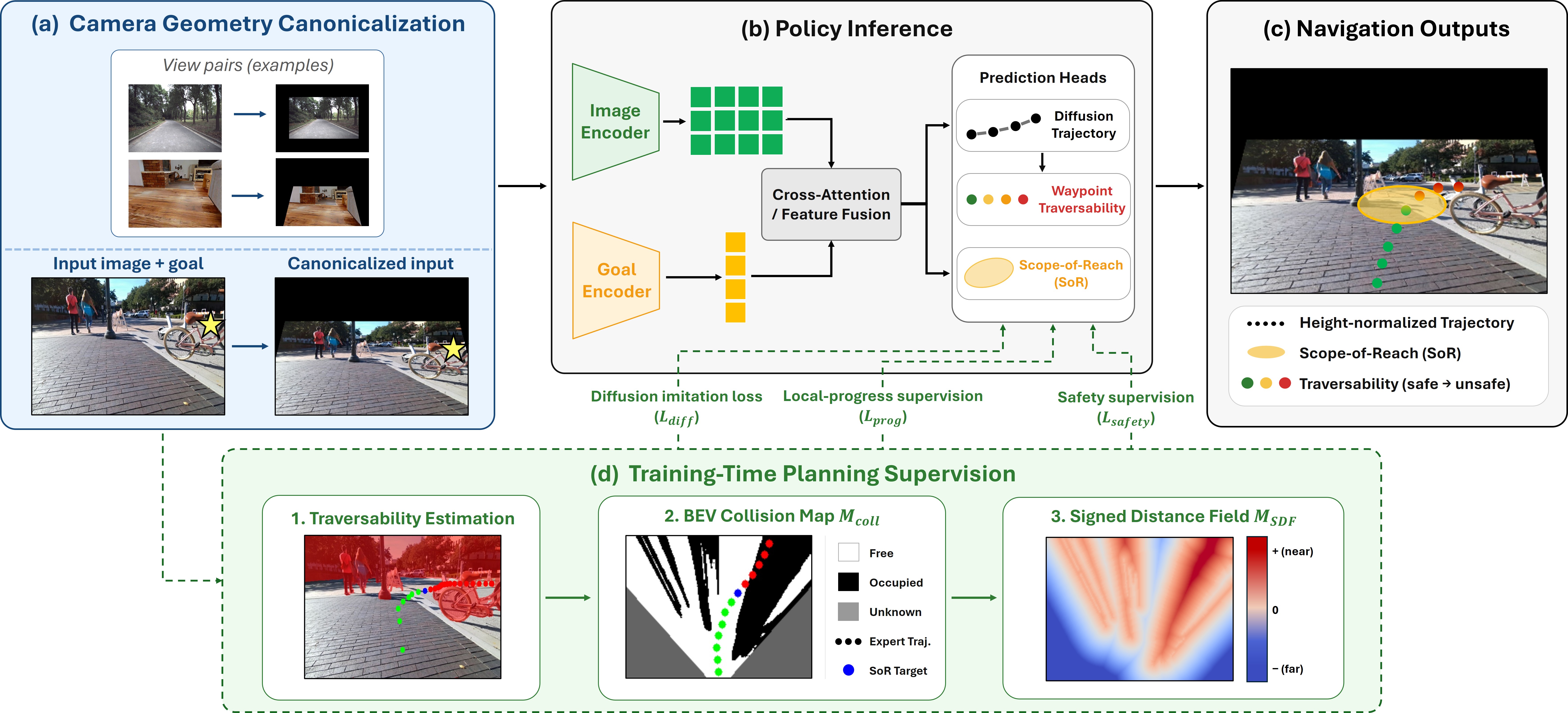}
    \caption{Overview of CanonNav. CanonNav canonicalizes the image and goal into a camera-consistent representation space. The network fuses image and goal features through cross-attention to predict height-normalized trajectories, per-waypoint traversability, and a Scope-of-Reach. During training, safety pseudo-labels and Scope-of-Reach targets from offline traversability estimation provide safety and local-progress supervision that complements trajectory imitation.}
    \label{fig:overall}
\end{figure*}

\section{Related Work}
\subsection{Visual Navigation Policy Learning}
End-to-end visual navigation predicts navigation behavior directly from visual observations. A major research direction trains policies via imitation learning on demonstrations collected across diverse environments and platforms~\citep{shah2021rapid, karnan2022socially, hirose2023sacson, hirose2025learning}. Representative methods such as GNM~\citep{shah2023gnm} and ViNT~\citep{shah2023vint} scale this paradigm to heterogeneous navigation data. Inspired by Diffusion Policy~\citep{chi2025diffusion}, recent methods employ generative planners that sample multiple trajectories~\citep{sridhar2024nomad, ren2025prior, son2026ordinal}. 

To complement imitation learning, prior work incorporates planning signals for safe and goal-directed navigation. iPlanner~\citep{Yang-RSS-23} learns a local planner with differentiable costmap objectives, and ViPlanner~\citep{roth2024viplanner} extends this framework with semantic traversability costs. LiMo~\citep{inglin2026morescalablevisualnavigation} augments demonstrations with planner-generated trajectories. Diffusion-based methods use such signals for trajectory generation or selection: NavDP~\citep{cai2025navdp} learns critic-based evaluation from privileged simulation supervision, while NaviDiffusor~\citep{zeng2025navidiffusor} guides diffusion sampling using differentiable goal and collision costs. Despite these advances, prior methods do not explicitly account for platform-dependent camera geometry when learning from cross-platform demonstrations. In contrast, CanonNav defines safety and local-progress supervision within a camera-consistent representation space, allowing these signals to complement imitation learning consistently across platforms. 

\subsection{Cross-Platform Visual Navigation}
Beyond dataset aggregation, recent work has addressed platform variation in visual navigation. X-Nav~\citep{wang2025x} trains embodiment-specific policies and distills them into a unified policy. Camera-view variation is typically addressed through augmentation or randomization. ExAug~\citep{hirose2023exaug} synthesizes counterfactual viewpoints from reconstructed 3D geometry, while RoVi-Aug~\citep{chen2024roviaug} uses generative augmentation to create demonstrations under different robot-camera viewpoints. RING~\citep{eftekhar2024one} randomizes embodiment and camera configurations in simulation to expose policies to diverse platform settings. While these methods broaden the training distribution over platform and viewpoint variations, they do not fundamentally resolve camera-dependent correspondence, limiting generalization to the variations covered during training.

\section{Method}
\subsection{Overall Framework}
    We consider point-goal visual navigation, where the agent predicts a robot-centric trajectory $\tau_t=\{(x_i,y_i)\}_{i=1}^{N}$ from an RGB observation $I_t$ and a relative goal $g_t=(r_t,\theta_t)$, with $x$ and $y$ denoting the forward and left directions. 

    The overall framework is illustrated in Fig.~\ref{fig:overall}. We first canonicalize the image and goal into $\tilde{I}_t$ and $\tilde{g}_t$ (Sec.~\ref{subsection:Canonicalization}). Image and goal encoders then extract an image feature $f_I \in \mathbb{R}^{H_p \times W_p \times C_I}$ and a goal feature $f_G \in \mathbb{R}^{C_G}$. We apply cross-attention using $f_G$ as the query and $f_I$ as the key and value, yielding a goal-conditioned visual feature $f_A \in \mathbb{R}^{C_G}$. The features are concatenated into the conditioning vector,
    \begin{equation}
    \label{eq:diffusion_feature}
        f_{\text{cond}} = [f_I^{\text{pool}}, f_G, f_A],
    \end{equation}
    where $f_I^{\text{pool}} \in \mathbb{R}^{C_I}$ is obtained by applying global average pooling to $f_I$. The conditioning vector $f_{\text{cond}}$ is used by two prediction branches: a diffusion policy predicts $B$ height-normalized trajectory candidates $\tilde{\tau}_t \in \mathbb{R}^{B \times N \times 2}$, while the Scope-of-Reach branch estimates a height-normalized local-progress region $(\tilde{\mu}_t, \tilde{\sigma}_t^2)$ (Sec.~\ref{subsection:auxiliary}). For each trajectory, a lightweight traversability network predicts per-waypoint traversability scores $\hat{q}_t \in \mathbb{R}^{B \times N}$ from waypoint coordinates and their corresponding image features. 
    
\subsection{Camera Geometry Canonicalization}
\label{subsection:Canonicalization}
    Most prior methods learn a policy $p(\tau_t \mid I_t, g_t)$ directly from image-goal-trajectory pairs. However, the underlying mapping also depends on the camera geometry $\phi=(K,R,t_\text{cam})$, which determines how trajectories correspond to the visual observation. Consequently, a policy conditioned only on $(I_t,g_t)$ must account for the mappings induced by varying camera geometries within a single conditional distribution:
    \begin{equation}
    p(\tau_t \mid I_t, g_t) = \int p(\tau_t \mid I_t, g_t, \phi) \, p(\phi \mid I_t) \, d\phi .
    \end{equation}    
    The policy must therefore implicitly infer the projection geometry from visual observations, an inherently ill-posed task that hinders learning from cross-platform demonstrations.

    To address this, we introduce \emph{camera geometry canonicalization}, which maps images, goals, and trajectories into a camera-consistent representation space. Assuming a front-centered camera, we reduce the extrinsics to pitch rotation $R_{\text{pitch}}$ and height $h$, giving $\phi=(K,R_{\text{pitch}},h)$. Given a canonical intrinsic matrix $\tilde{K}$ parameterized by $(\tilde{f}_x,\tilde{f}_y,\tilde{c}_x,\tilde{c}_y)$, we transform $I_t$ into a canonical front-facing view $\tilde{I}_t$:
    \begin{equation}
    \label{eq:canonicalization}
    (\tilde{u}, \tilde{v}, 1)^\top \propto 
    \tilde{K} R_{\text{pitch}}^{-1} K^{-1} (u, v, 1)^\top,
    \end{equation}
    where $(u,v)$ and $(\tilde{u},\tilde{v})$ denote the original and transformed pixel coordinates. This canonical view removes variation due to camera intrinsics and pitch rotation.

    Under the canonical front-facing camera, a ground-plane point $(x,y)$ is represented in the camera frame as $(-y,-h,x)^\top$, whose projection is
    \begin{equation}
    \label{eq:projection}
    \tilde{u} = \tilde{c}_x - \tilde{f}_x \frac{y}{x},
    \qquad
    \tilde{v} = \tilde{c}_y - \tilde{f}_y \frac{h}{x}.
    \end{equation}
    Since the projection depends only on the ratios $y/x$ and $h/x$, we normalize the trajectory by camera height as $(\tilde{x},\tilde{y})=(x/h,y/h)$, which preserves the projection while eliminating explicit dependence on $h$:
    \begin{equation}
    \label{eq:normalized_projection}
    \tilde{u} = \tilde{c}_x - \tilde{f}_x \frac{\tilde{y}}{\tilde{x}},
    \qquad
    \tilde{v} = \tilde{c}_y - \tilde{f}_y \frac{1}{\tilde{x}}.
    \end{equation}
    We also normalize the goal as $\tilde{g}_t=(r_t/h,\theta_t)$ to align it with the same canonical space. Together, predicting the height-normalized trajectory $\tilde{\tau}_t$ from canonicalized inputs $\tilde{I}_t$ and $\tilde{g}_t$ enables consistent policy learning by disentangling navigation behavior from platform-dependent camera geometry.

\subsection{Planning Supervision in the Canonical Space}
    \label{subsection:auxiliary}
    For this section, we omit the time index $t$, as all quantities refer to a single timestep. CanonNav complements diffusion-based imitation learning with safety and local-progress supervision defined within the canonical representation.

    \paragraph{Diffusion-based imitation learning}    
    Following prior work~\citep{sridhar2024nomad}, we use a diffusion policy~\citep{chi2025diffusion} for trajectory generation. A height-normalized expert trajectory $\tilde{\tau}^{\ast}$ is perturbed with Gaussian noise at a sampled diffusion timestep $k$, and the denoising network $\epsilon_{\theta}$ predicts the noise conditioned on $f_{\mathrm{cond}}$. The diffusion objective $\mathcal{L}_{\mathrm{diff}}$ is the standard DDPM loss between the sampled and predicted noise~\citep{ho2020denoising}.

    While $\mathcal{L}_{\mathrm{diff}}$ learns to reproduce the demonstrated trajectory distribution, it does not explicitly indicate which regions are safe or where the robot should advance. We therefore introduce safety and local-progress supervision signals that provide explicit guidance to the denoised trajectory estimate. Given a noisy trajectory $\tilde{\tau}_k$, we recover the clean height-normalized estimate as
    \begin{equation}
    \label{eq:denoised_trajectory}
    \hat{\tau}_0
    =
    \frac{
        \tilde{\tau}_k
        -
        \sqrt{1-\bar{\alpha}_k}\,
        \epsilon_\theta(\tilde{\tau}_k,k,f_{\text{cond}})
    }{
        \sqrt{\bar{\alpha}_k}
    },
    \end{equation}
    where $\bar{\alpha}_k$ is the cumulative diffusion coefficient defined by the DDPM noise schedule. The subsequent supervision objectives are formulated based on $\hat{\tau}_0$.
    
    \paragraph{Safety supervision}
    We use two safety objectives: a trajectory-level collision loss that directly shapes trajectory generation and a waypoint-level traversability loss for execution-time path selection:
    \begin{equation}
        \mathcal{L}_{\mathrm{safety}}
        =
        \lambda_{\mathrm{coll}}\mathcal{L}_{\mathrm{coll}}
        +
        \lambda_{\mathrm{trav}}\mathcal{L}_{\mathrm{trav}}.
    \end{equation}

    Safety supervision is derived from the BEV-based pipeline as shown in Fig.~\ref{fig:overall}(d). First, an offline traversability estimator predicts a binary traversability mask for each training image. Using the known camera geometry $\phi$, we project the mask onto the local ground plane to obtain a pseudo-BEV collision map $M_{\mathrm{coll}}$, which is then converted into a metric signed distance field $M_{\mathrm{sdf}}$. Note that neither traversability estimation nor pseudo-label generation is required at inference.

    The trajectory-level collision loss encourages the generated path to avoid unsafe regions. For each height-normalized waypoint $\hat{p}_i$ in $\hat{\tau}_0$, we sample the SDF value at its metric position $h\hat{p}_i$ and penalize clearance below the safety margin:
    \begin{equation}
    \begin{aligned}
        c_i
        &=
        \mathrm{softplus}
        \left(
        \frac{
            d_{\mathrm{safe}}-M_{\mathrm{sdf}}(h\hat{p}_i)
        }{\eta}
        \right), \\
        \mathcal{L}_{\mathrm{coll}}
        &=
        \frac{1}{\gamma}
        \log
        \sum_{i=1}^{N}
        \exp(\gamma c_i),
    \end{aligned}
    \end{equation}
    where $d_{\mathrm{safe}}$ denotes the safety margin, and $\eta$ and $\gamma$ control the sharpness of the penalty.
    
    We additionally train a lightweight traversability estimator for trajectory selection. For each waypoint $\hat{p}_i$, we bilinearly sample the image feature $f_I$ at its image projection. The estimator takes the sampled feature and waypoint coordinate as inputs and predicts a traversability score $\hat{q}_i \in [0,1]$:
    \begin{equation}
        \mathcal{L}_{\mathrm{trav}}
        =
        \frac{1}{N}
        \sum_{i=1}^{N}
        \mathrm{BCE}(\hat q_i,q_i^\ast).
    \end{equation}
    We stop gradients through $\hat{p}_i$ and the sampled image feature, so $\mathcal{L}_{\mathrm{trav}}$ updates only the traversability network. For waypoints within the field of view, we set the target $q_i^\ast=1$ when $M_{\mathrm{sdf}}(h\hat{p}_i)>d_{\mathrm{robot}}$ and $0$ otherwise, where $d_{\mathrm{robot}}$ denotes the robot size. For out-of-view waypoints, we set $q_i^\ast=0.5$.

    \paragraph{Local-progress supervision}
    In cluttered scenes where the goal cannot be reached directly, the policy should identify a local region that enables feasible progress. To represent this intermediate planning decision, we define the \emph{Scope-of-Reach} (SoR) as a locally appropriate region indicating where the robot should advance, as illustrated in Fig.~\ref{fig:overall}(c). We model the SoR as a BEV Gaussian distribution:
    \begin{equation}
        \mathcal{S} = \mathcal{N}
        \left(
            \mu,
            \mathrm{diag}\left(\sigma^2\right)
        \right).
    \end{equation}
    The SoR prediction branch takes $f_{\text{cond}}$ as input and predicts height-normalized parameters $(\tilde{\mu}, \tilde{\sigma})=(\mu/h, \sigma/h)$. 

    For each training sample, we derive  a target point $s^\ast$ from the expert trajectory $\tau^\ast=\{p_i^\ast\}_{i=1}^{N}$ and the collision map $M_{\mathrm{coll}}$. As shown in Fig.~\ref{fig:overall}(d), we select $s^\ast$ as the last waypoint before the trajectory first enters a collision region or leaves the camera field of view. If the trajectory remains visible and collision-free until its final waypoint, we set $s^\ast=p_N^\ast$. The local-progress supervision uses this target to learn a safe intermediate region and align trajectory generation with it:
    \begin{equation}
        \mathcal{L}_{\mathrm{prog}}
        =
        \lambda_{\mathrm{pred}}
        \mathcal{L}_{\mathrm{pred}}
        +
        \lambda_{\mathrm{neg}}
        \mathcal{L}_{\mathrm{neg}}
        +
        \lambda_{\mathrm{cons}}
        \mathcal{L}_{\mathrm{cons}}.
    \end{equation}

    We first fit the predicted SoR to the height-normalized target $\tilde{s}^\ast=s^\ast/h$ using Gaussian negative log-likelihood and Smooth L1 losses:
    \begin{equation}
        \mathcal{L}_{\mathrm{pred}}
        =
        -\log
        \mathcal{N}
        \left(
            \tilde{s}^\ast ;
            \tilde{\mu},
            \mathrm{diag}\left(\tilde{\sigma}^2\right)
        \right)
        +
        \mathrm{SmoothL1}
        \left(
            \tilde{\mu},
            \tilde{s}^\ast
        \right).
    \end{equation}

    Since target fitting alone does not ensure that the predicted SoR avoids unsafe regions, we penalize its overlap with collision cells in the BEV collision map $M_{\mathrm{coll}}$, where 1 denotes collision and 0 denotes free space:
    \begin{equation}
        \mathcal{L}_{\mathrm{neg}}
        =
        \int
        M_{\mathrm{coll}}(x,y)\,\mathcal{S}(x,y)\, dx\,dy.
    \end{equation}
    In practice, we approximate the integral by summing the discretized SoR probability mass over collision cells.

    Finally, we enforce consistency between the generated trajectory and the predicted SoR to guide goal-directed local progress. We compute the Mahalanobis distance from each denoised waypoint $\hat{p}_i$ to the SoR and apply a soft-min objective that encourages at least one waypoint to pass near the predicted region:
    \begin{equation}
    \begin{aligned}
        d_i
        &=
        (\hat{p}_i-\tilde{\mu})^\top
        \mathrm{diag}(\tilde{\sigma}^2)^{-1}
        (\hat{p}_i-\tilde{\mu}),
        \\
        \mathcal{L}_{\mathrm{cons}}
        &=
        -\rho
        \log
        \sum_{i=1}^{N}
        \exp
        \left(
            -\frac{d_i}{\rho}
        \right),
    \end{aligned}
    \end{equation}
    where $\rho$ controls the soft-min smoothness. 

\begin{table*}[t]
    \centering
    \small
    \setlength{\tabcolsep}{3pt}
    \renewcommand{\arraystretch}{1.15}
    \begin{tabular}{l|cccccc|cccccc}
    \toprule
    & \multicolumn{6}{c|}{\textbf{CitySim (Outdoor)}} & \multicolumn{6}{c}{\textbf{AWS Hospital (Indoor)}} \\
    \cmidrule(lr){2-7} \cmidrule(lr){8-13}
    \multicolumn{1}{c|}{\textbf{Method}} & \multicolumn{2}{c}{6m} & \multicolumn{2}{c}{12m} & \multicolumn{2}{c|}{20m}
    & \multicolumn{2}{c}{6m} & \multicolumn{2}{c}{9m} & \multicolumn{2}{c}{12m} \\
    \cmidrule(lr){2-3} \cmidrule(lr){4-5} \cmidrule(lr){6-7}
    \cmidrule(lr){8-9} \cmidrule(lr){10-11} \cmidrule(lr){12-13}
    & SR $\uparrow$ & Coll. $\downarrow$
    & SR $\uparrow$ & Coll. $\downarrow$
    & SR $\uparrow$ & Coll. $\downarrow$
    & SR $\uparrow$ & Coll. $\downarrow$
    & SR $\uparrow$ & Coll. $\downarrow$
    & SR $\uparrow$ & Coll. $\downarrow$ \\
    \midrule
    ViNT~\citep{shah2023vint}      & 93.8 & 0.69 & 63.0 & 4.27 & 18.6 & 8.03 & 62.1 & 9.23 & 38.8 & 13.07 & 30.0 & 13.10 \\
    NoMaD~\citep{sridhar2024nomad} & 98.0 & 0.43 & 61.6 & 4.29 & 24.5 & 7.00 & 65.6 & 8.15 & 40.2 & 12.22& 29.8 & 13.30 \\
    OmniVLA-edge~\citep{hirose2025omnivla} & 89.4 & 0.76 & 61.4 & 2.87 & 28.5 & 4.26 & 72.2 & 6.11 & 41.7 & 10.98 & 31.1 & 11.68 \\
    LiMo~\citep{inglin2026morescalablevisualnavigation} & 99.1 & 0.26 & 79.6 & 2.16 & 48.3 & 4.25 & 81.0 & 4.17 & 54.3 & 8.03 & 39.7 & 9.92 \\
    ViPlanner$^\ast$~\citep{roth2024viplanner} & 81.6 & 0.61 & 62.9 & 3.00 & 39.3 & 4.87 & 86.6 & 2.09 & 69.3 & 4.23 & 52.2 & 6.04 \\
    NavDP$^\ast$~\citep{cai2025navdp} & \textbf{99.7} & \textbf{0.07} & \underline{93.7} & \underline{0.54} & \underline{71.2} & \underline{2.21} & \textbf{98.3} & \textbf{0.53} & \textbf{93.1} & \textbf{1.22} & \underline{82.9} & \textbf{1.91} \\
    \midrule
    CanonNav (Ours) & \underline{99.4} & \underline{0.20} & \textbf{94.1} & \textbf{0.53} & \textbf{86.4} & \textbf{0.89} & \underline{97.1} & \underline{1.26} & \underline{92.2} & \underline{1.77} & \textbf{85.5} & \underline{2.37} \\
    \bottomrule
    \end{tabular}
    \caption{Quantitative comparison across environments and subgoal intervals, averaged over camera configurations. Methods with $^\ast$ use RGB-D input, while others use RGB only. Best and second-best results are shown in bold and underlined, respectively.}
    \label{tab:main_comparison}
\end{table*}

    \paragraph{Overall training objective.}
    To avoid unrealistic trajectories, we apply a smoothness regularizer $\mathcal{L}_{\mathrm{smooth}}$ that penalizes large second-order differences between consecutive waypoints. We jointly optimize all four objectives as
    \begin{equation}
    \mathcal{L}
    =
    \lambda_{\mathrm{diff}}\mathcal{L}_{\mathrm{diff}}
    +
    \mathcal{L}_{\mathrm{safety}}
    +
    \mathcal{L}_{\mathrm{prog}}
    +
    \lambda_{\mathrm{smooth}}\mathcal{L}_{\mathrm{smooth}}.
    \end{equation}

\subsection{Inference and Implementation Details}
    \paragraph{Trajectory selection}
    At inference, we select a trajectory from the candidate set using the waypoint traversability scores. Each score $\hat{q}_{i}$ is interpreted as a unit-length survival probability for the path segment from $p_{i-1}$ to $p_i$. Accumulating these survival probabilities along the path yields the expected reach point $\bar{p}$ via arc-length interpolation. Each candidate is evaluated by three criteria: goal progress, measured by the reduction in goal distance at $\bar{p}$; displacement, which captures how far $\bar{p}$ is from the current position; and temporal consistency with the previously selected trajectory. We execute the candidate with the highest combined score.

    \paragraph{Implementation details}
    We train CanonNav on public navigation datasets SCAND~\citep{karnan2022socially}, HuRoN~\citep{hirose2023sacson}, and BotanicGarden~\citep{liu2024botanicgarden}, together with our real-world Husky data and simulated HM3D data~\citep{ramakrishnan2021hm3d}. After preprocessing, we obtain 8.35M image-goal-trajectory pairs for training. We use ViTA~\citep{hwang2026general} for offline traversability estimation during preprocessing. The image encoder is a pretrained DINOv3~\citep{simeoni2025dinov3} ViT-S+ backbone adapted with LoRA~\citep{hu2022lora}. Images are canonicalized to $336\times256$ using canonical intrinsics $\tilde{f}_x=\tilde{f}_y=\tilde{c}_x=168$ and $\tilde{c}_y=128$. Further details are provided in the supplementary material.

\section{Experiments}
\subsection{Simulation Experimental Setup}
    We evaluate CanonNav in two simulation environments for visual navigation: Gazebo CitySim~\citep{koenig2004design} for outdoor navigation and AWS Hospital~\citep{aws_robomaker_hospital_world} for indoor navigation. We conduct extensive experiments to answer the following questions:
    \begin{enumerate}[label=\textbf{Q\arabic*.}, leftmargin=3.0em, itemsep=0em, topsep=0.1em]
        \item Does camera geometry canonicalization improve robustness across varying camera configurations?
        \item Do safety and local-progress supervision improve navigation performance in scenarios requiring complex local planning?
    \end{enumerate}

    To evaluate \textbf{Q1}, we compare navigation performance across camera configurations with varying horizontal fields of view ($70^\circ$, $80^\circ$, and $90^\circ$), pitch angles ($0^\circ$ front-facing and $10^\circ$, $20^\circ$ downward), and camera heights (0.55 m, 0.75 m for CitySim, and 0.45 m, 0.65 m for AWS Hospital), resulting in 18 different camera configurations for each environment.

    For \textbf{Q2}, we vary the subgoal distance to evaluate different levels of local-planning difficulty. We define 10 reference paths in each environment and partition them at fixed intervals to obtain subgoals at 6 m, 12 m, and 20 m in CitySim, and 6 m, 9 m, and 12 m in AWS Hospital. If the agent fails to reach a subgoal, it is respawned at that subgoal and continues to the next one. This produces 312 evaluation cases in CitySim and 264 in AWS Hospital. Each case is evaluated under all 18 camera configurations and repeated 10 times, resulting in 56,160 and 47,520 rollouts, respectively.

\begin{figure}[!b]
    \centering
    \includegraphics[width=\linewidth]{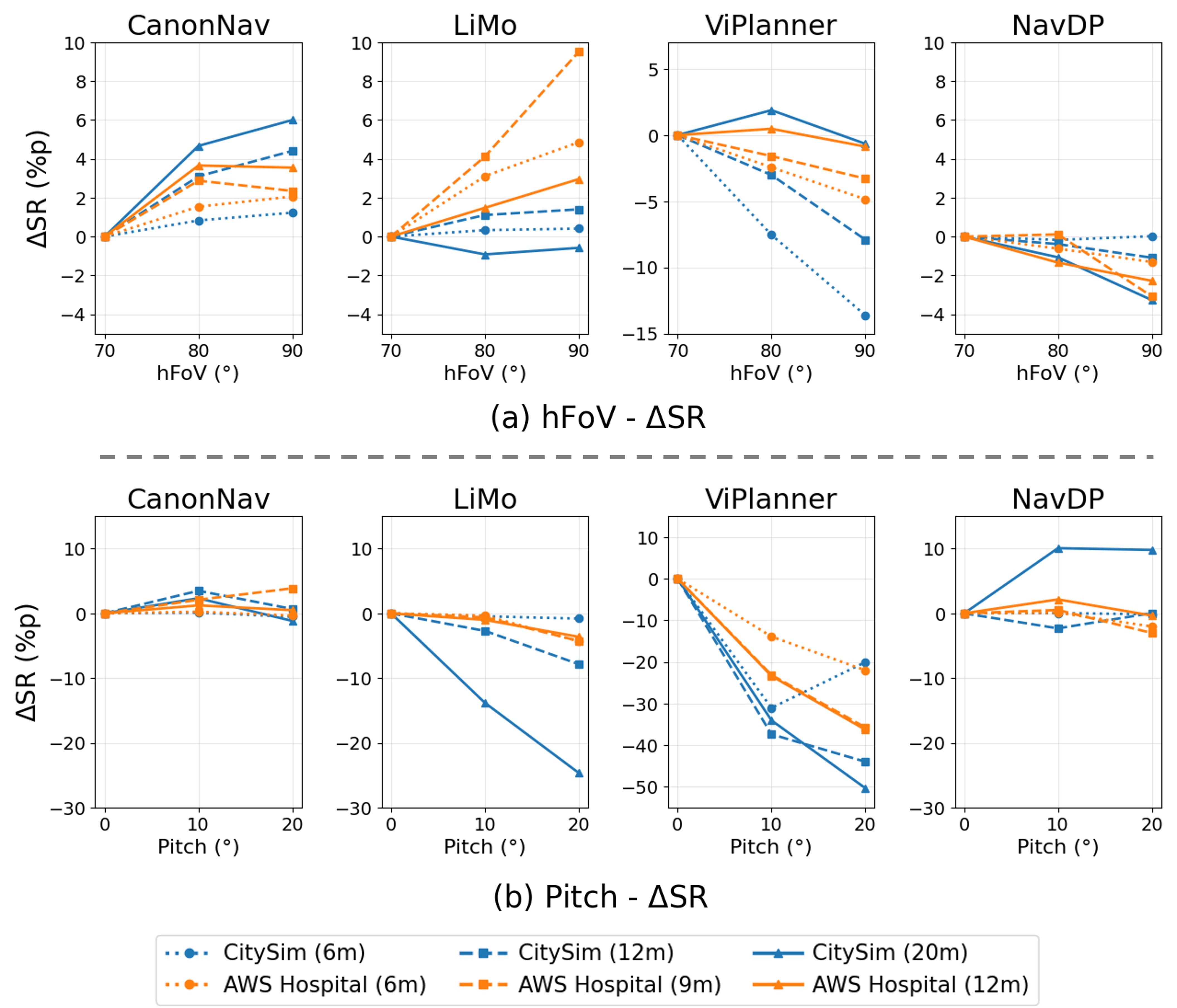}
    \caption{Changes in success rate ($\Delta\mathrm{SR}$) with respect to (a) horizontal FoV and (b) camera pitch. CanonNav's performance improves with wider horizontal FoVs and remains robust to pitch changes. Note that the ViPlanner subplots use different y-axis ranges for visualization.}
    \label{fig:ex_fov_pitch}
\end{figure}

\begin{figure*}[!t]
    \centering
    \includegraphics[width=\textwidth]{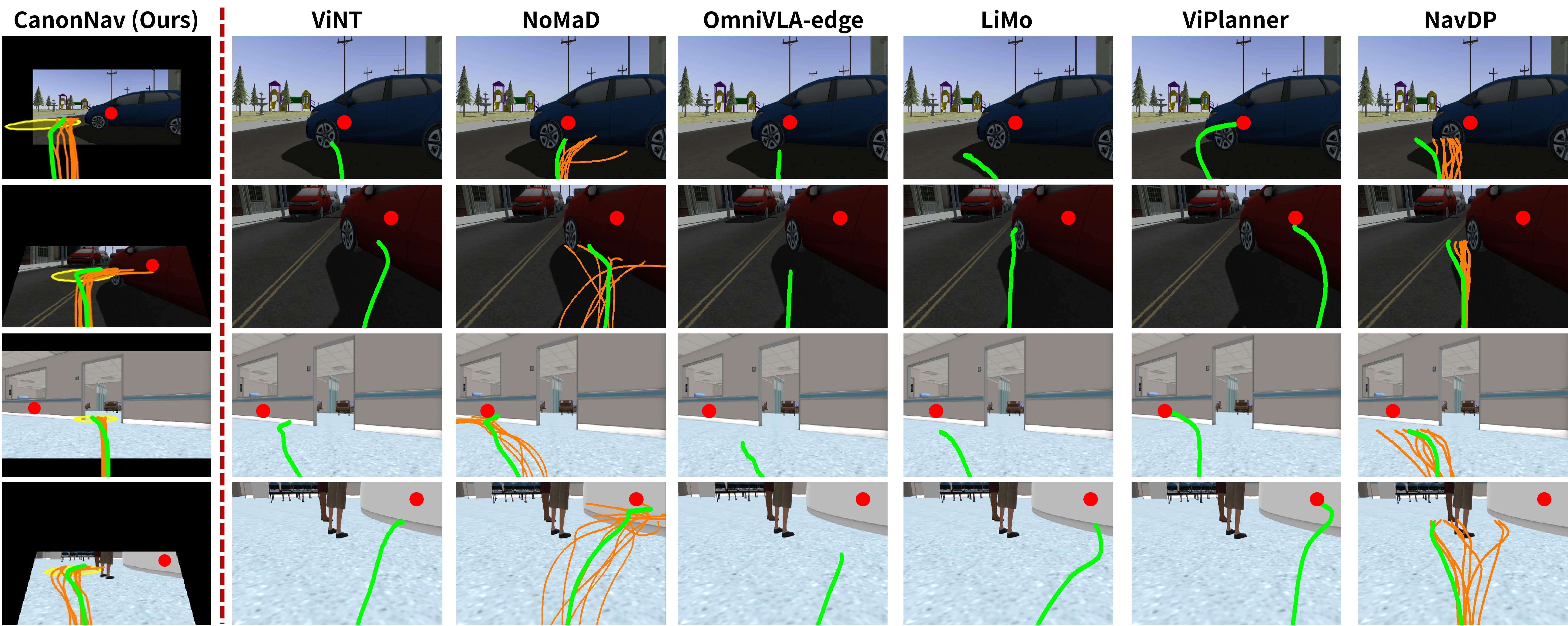}
    \caption{Qualitative results in simulation. The red point indicates the goal, with the direct path blocked by obstacles. The green curve shows the executed trajectory, while candidate trajectories are shown in orange for methods that generate multiple trajectory hypotheses. CanonNav performs robust obstacle-avoiding navigation across different camera configurations by canonicalizing diverse camera views and predicting the Scope-of-Reach (yellow region).}
    \label{fig:experiments_qualitative}
\end{figure*}

\subsection{Baselines and Evaluation Metrics}
    We compare CanonNav with six baselines: four RGB-based methods, ViNT~\citep{shah2023vint}, NoMaD~\citep{sridhar2024nomad}, OmniVLA~\citep{hirose2025omnivla}, and LiMo~\citep{inglin2026morescalablevisualnavigation}, and two RGB-D-based methods, ViPlanner~\citep{roth2024viplanner} and NavDP~\citep{cai2025navdp}. Since ViNT and NoMaD were originally designed for image-goal navigation, we retrain them on our dataset to support point-goal navigation. For the remaining baselines, we use publicly released checkpoints. For OmniVLA, we evaluate OmniVLA-edge, as it outperformed the original OmniVLA in our setting. All methods are evaluated zero-shot in environments unseen during training. We report subgoal-level success rate (\textbf{SR}, $\%$) and collisions per 100\,m traveled (\textbf{Coll.}).

\subsection{Simulation Results}
    Table~\ref{tab:main_comparison} reports navigation performance averaged over all camera configurations. CanonNav achieves the highest success rate at the longest subgoal interval in both environments, reaching 86.4\% at 20\,m in CitySim and 85.5\% at 12\,m in AWS Hospital. It also remains competitive with the depth-based NavDP at shorter intervals despite using only RGB.

    \paragraph{Robustness to camera variation (\textbf{Q1}).}
    Fig.~\ref{fig:ex_fov_pitch} reports the change in success rate ($\Delta\mathrm{SR}$) under varying hFoVs and camera pitches relative to the reference settings of $70^\circ$ hFoV and $0^\circ$ pitch. As a wider hFoV provides more scene context, it improves navigation when the additional information is effectively exploited. CanonNav and LiMo generally benefit from wider views, whereas ViPlanner and NavDP degrade as hFoV increases. Pitch variation substantially alters the observed viewpoint, with its largest effect observed in the CitySim 20\,m setting. LiMo and ViPlanner show substantial SR drops of 24.66 and 50.34 percentage points (pp), respectively, while NavDP improves by 9.80\,pp, indicating sensitivity to the viewpoint change. In contrast, CanonNav varies by only 1.17\,pp in the same setting, and its largest pitch-induced change remains limited to 3.87\,pp at 9\,m in AWS Hospital. Among the compared methods, only CanonNav benefits from wider hFoVs while maintaining stable performance under pitch variation, indicating that camera geometry canonicalization reduces sensitivity to camera changes.

    \paragraph{Effectiveness of safety and local-progress supervision (\textbf{Q2}).}
    Fig.~\ref{fig:experiments_qualitative} compares navigation behaviors in both environments. Without explicit planning supervision, RGB baselines tend to move directly toward the goal, which can lead to collisions when obstacles block the route. ViPlanner and NavDP exploit depth cues for obstacle avoidance, while NavDP tends to react at close range rather than plan detours in advance. In contrast, CanonNav plans earlier detours by predicting the SoR and generating collision-aware trajectories toward the predicted region. This behavior is reflected in Table~\ref{tab:main_comparison}: CanonNav achieves the highest success rate at the longest subgoal interval in both environments while maintaining low collision rates using only RGB. At 20\,m in CitySim, the longest interval evaluated, it reduces collisions from 2.21 to 0.89 per 100\,m compared with NavDP. In AWS Hospital, it remains competitive in collision rate while achieving the highest success rate at 12\,m. These results suggest that safety and local-progress supervision improve collision-aware motion and effective local planning in challenging scenarios.

\begin{figure*}[!t]
    \centering
    \includegraphics[width=\textwidth]{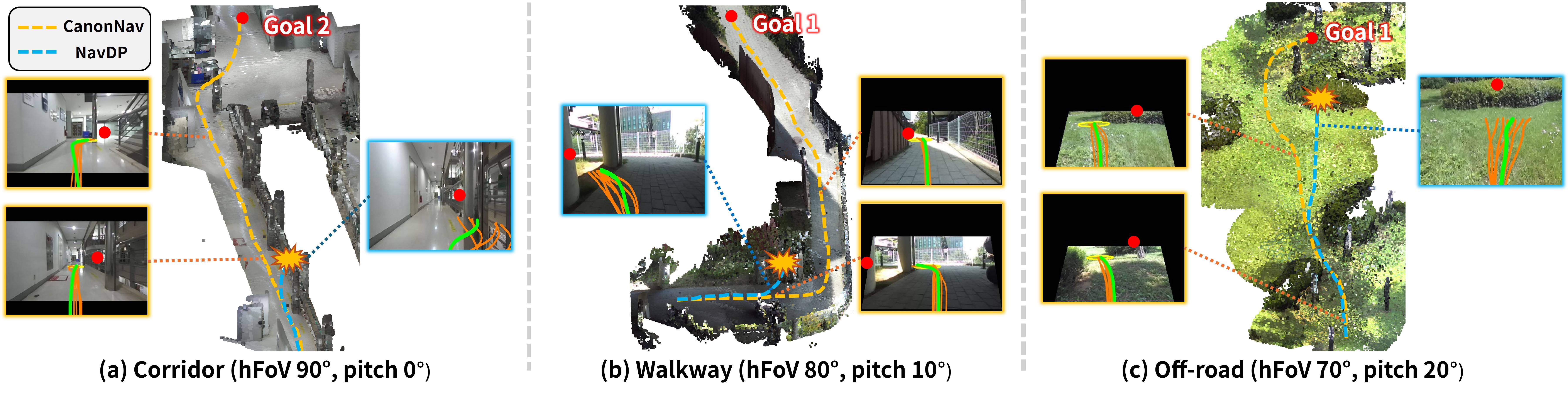}
    \caption{Qualitative comparison in real-world experiments. The images show each method's planning results, while the dashed lines indicate the executed trajectories. NavDP relies on short-range obstacle avoidance from depth cues and fails to plan toward the distant goal, whereas CanonNav generates collision-aware trajectories that enable successful progress toward it.}
    \label{fig:real_worlds}
\end{figure*}

\subsection{Ablation Study}
    We ablate the three main components of CanonNav: camera geometry canonicalization, safety supervision, and local-progress supervision. Without canonicalization, images are directly resized to the network input resolution, while goals and trajectories remain in metric coordinates.

    Table~\ref{tab:ablation_components} summarizes the results. Under default imitation-only training, we found that the pretrained DINOv3 adapted with LoRA substantially outperforms the model trained from scratch. Adding camera geometry canonicalization to this model improves performance in both environments, showing that a camera-consistent representation benefits imitation learning even without planning supervision. With the canonicalized imitation model, applying either safety or local-progress supervision substantially improves performance, while combining both achieves the best results. When both supervision objectives are applied without canonicalization, SR drops to 77.1\% and 75.2\%, respectively, while the standard deviations increase from 3.8 to 7.7 and from 3.9 to 7.6. This suggests that a camera-consistent representation space is critical for safety and local-progress supervision to remain effective across varying camera configurations.

\begin{table}[t]
    \centering
    \small
    \setlength{\tabcolsep}{3pt}
    \renewcommand{\arraystretch}{1.15}

    \begin{tabular}{
        @{}l
        >{\centering\arraybackslash}p{1.35cm}
        >{\centering\arraybackslash}p{1.6cm}
        |
        >{\centering\arraybackslash}p{1.35cm}
        >{\centering\arraybackslash}p{1.7cm}
        @{}
    }
    \toprule
    \textbf{Setting}
    & \multicolumn{2}{c|}{\textbf{CitySim (20m)}}
    & \multicolumn{2}{c}{\textbf{AWS Hospital (12m)}} \\
    \cmidrule(lr){2-3} \cmidrule(lr){4-5}

    & SR $\uparrow$ & Coll. $\downarrow$
    & SR $\uparrow$ & Coll. $\downarrow$ \\
    \midrule

    -- (Scratch)
    & 28.8 $\pm$ \textbf{2.1} & 6.36  $\pm$ 0.39
    & 32.4 $\pm$ \textbf{2.0} & 12.15 $\pm$ 0.60 \\

    -- (LoRA)
    & 62.6 $\pm$ 6.6 & 2.99  $\pm$ 0.52
    & 67.0 $\pm$ 5.1  & 4.76 $\pm$ 0.78  \\

    C
    & 65.6 $\pm$ 5.6 &  2.52 $\pm$ 0.45
    & 70.0 $\pm$ 6.3 & 4.62 $\pm$ 0.91 \\

    C+S
    & 83.0 $\pm$ 3.3 & 1.15 $\pm$ 0.23
    & 79.8 $\pm$ 5.5 & 3.42 $\pm$ 0.79 \\

    C+P
    & 79.1 $\pm$ 6.5 & 1.53 $\pm$ 0.36
    & 81.2 $\pm$ 5.4 & 3.13 $\pm$ 0.72 \\

    S+P
    & 77.1 $\pm$ 7.7  & 1.66 $\pm$ 0.66
    & 75.2 $\pm$ 7.6 & 4.04 $\pm$ 0.96 \\
    \midrule

    \textbf{C+S+P}
    & \textbf{86.4} $\pm$ 3.8 & \textbf{0.89} $\pm$ \textbf{0.19}
    & \textbf{85.5} $\pm$ 3.9 & \textbf{2.37} $\pm$ \textbf{0.54} \\
    \bottomrule
    \end{tabular}

    \vspace{2pt}
    \caption{Component ablation for CanonNav. C, S, and P denote camera geometry canonicalization, safety supervision, and local-progress supervision, respectively.}
    \label{tab:ablation_components}
\end{table}

\begin{table}[t]
    \centering
    \small
    \setlength{\tabcolsep}{3.2pt}
    \renewcommand{\arraystretch}{1.1}
    \begin{tabular}{llcc|cc}
        \toprule
        \multirow{2}{*}{\textbf{Environment}}
        & \multirow{2}{*}{\textbf{Subgoal}}
        & \multicolumn{2}{c|}{\textbf{NavDP}}
        & \multicolumn{2}{c}{\textbf{CanonNav}} \\
        \cmidrule(lr){3-4}\cmidrule(lr){5-6}
        &
        & SR $\uparrow$ & Int. $\downarrow$
        & SR $\uparrow$ & Int. $\downarrow$ \\
        \midrule
        \multirow{2}{*}{Corridor}
        & G1 (25\,m) & 3 / 10 & 1.3 & {8 / 10} & {0.3}\\
        & G2 (41\,m) & 0 / 10 & 3.8 & {7 / 10} & {0.4} \\
        \midrule
        \multirow{2}{*}{Walkway}
        & G1 (32\,m) & 1 / 10 & 1.5 & {9 / 10} & {0.1} \\
        & G2 (33\,m) & 9 / 10 & 0.1 & {10 / 10} & {0.0} \\
        \midrule
        Off-road
        & G1 (27\,m) & 4 / 10 & 1.2 & {7 / 10} & {0.5} \\
        \midrule
        \textbf{Overall}
        & & 17 / 50 & 1.58 & \textbf{41 / 50} & \textbf{0.26} \\
        
        \bottomrule
    \end{tabular}
    \caption{Real-world navigation performance. SR reports trials completed without intervention, while Int. denotes the average interventions required to reach each subgoal.}
    \label{tab:real_world}
\end{table}

\subsection{Real-world Experiments}
    We deploy CanonNav and NavDP on a Clearpath Husky equipped with a ZED 2i stereo camera and a Jetson AGX Orin 64GB. CanonNav runs onboard at 4.5\,Hz. We evaluate five subgoals ranging from 25\,m to 41\,m across three environments, with 10 trials per subgoal. Each environment uses a different camera hFoV and pitch to reflect diverse deployment settings. We report the success rate and the average number of human interventions to reach each subgoal.
    
    Table~\ref{tab:real_world} and Fig.~\ref{fig:real_worlds} present the quantitative and qualitative results, respectively. CanonNav achieves higher success rates and requires fewer interventions than NavDP in all three environments.  The failure cases suggest that NavDP relies on local depth cues for short-range obstacle avoidance rather than broader visual context for planning. In Fig.~\ref{fig:real_worlds}(a,b), it steers toward gaps in the corridor fence and drifts off the walkway. In Fig.~\ref{fig:real_worlds}(c), it approaches the distant bush before attempting to avoid it. In contrast, CanonNav leverages broader visual context to infer an appropriate SoR and generate collision-aware trajectories toward it. Further qualitative results are provided in the supplementary material. Overall, these results demonstrate CanonNav's robustness across diverse real-world environments and camera settings.

\section{Conclusion}
In this work, we presented CanonNav, a framework for learning robust visual navigation from cross-platform demonstrations. Camera geometry canonicalization establishes a camera-consistent representation space that disentangles navigation behavior from platform-dependent camera geometry. Within this space, CanonNav derives safety and local-progress supervision from offline traversability estimates, enabling collision-aware and goal-directed planning. Extensive experiments show that, despite RGB-only inference, CanonNav consistently outperforms RGB-based baselines and even surpasses RGB-D-based methods in challenging scenarios. Nevertheless, CanonNav has two limitations. When the original camera has a wider field of view than the predefined canonical view, canonicalization may reduce visual context. Moreover, although CanonNav addresses variations in camera geometry, it does not model policy variations across robot embodiments. Future work will address these limitations.
\bigskip

\bibliography{aaai2027}

\input{Supplementary.tex}

\end{document}

%% file: Supplementary.tex
\setcounter{secnumdepth}{2} 
\setcounter{figure}{0}
\setcounter{table}{0}
\setcounter{equation}{0}
\setcounter{section}{0}
\renewcommand{\thesection}{S\arabic{section}}
\renewcommand{\thesubsection}{\thesection.\arabic{subsection}}
\renewcommand{\thefigure}{S\arabic{figure}}
\renewcommand{\thetable}{S\arabic{table}}
\renewcommand{\theequation}{S\arabic{equation}}

\twocolumn[
\begin{center}
    {\LARGE\bfseries
    CanonNav: Disentangling Navigation Behavior from Camera Geometry\\[1pt]
    in Cross-Platform Visual Navigation\\[5pt]
    Supplementary Material\par
    }
\end{center}
\vspace{40pt}
]


\section{Additional Method Details}
\subsection{Details of Path Selection}
\label{supp_section:path_selection}
\begin{figure}[t]
    \centering
    \includegraphics[width=\linewidth]{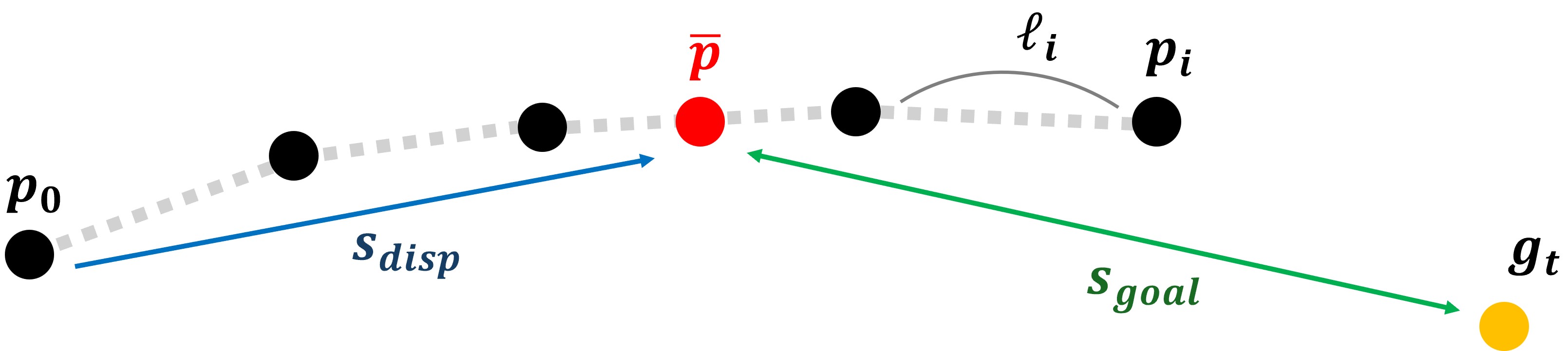}
    \caption{Candidate path selection.
    For each candidate path, traversability scores are used to estimate the expected reach point \(\bar{p}\). The candidate paths are then evaluated based on goal advance, displacement, and temporal consistency, and the highest-scoring path is selected.}
    \label{supp:path_selection}
\end{figure}

As illustrated in Fig.~\ref{supp:path_selection}, we select the final path from a set of candidates by first estimating the expected reach point \(\bar{p}\) for each candidate.

Given a candidate path
\[
\tau=\{p_0,p_1,\ldots,p_N\}, \qquad p_i\in\mathbb{R}^2,
\]
we define the length of the \(i\)-th segment as
\[
\ell_i=\|p_i-p_{i-1}\|_2, \qquad i=1,\ldots,N.
\]
For each segment from \(p_{i-1}\) to \(p_i\), the model outputs a traversability score \(q_i\in(0,1)\). We interpret \(q_i\) as the unit-distance survival probability along that segment, so that the probability of successfully traversing a segment of length \(\ell_i\) is \(q_i^{\ell_i}\).

The probability of reaching the beginning of segment \(i\) is the product of the success probabilities of all previous segments:
\[
R_i=\prod_{j=1}^{i-1} q_j^{\ell_j},
\]
where \(R_1=1\).
Conditioned on reaching segment \(i\), the expected travel distance within that segment is
\[
d_i
=
\int_0^{\ell_i} q_i^x \, dx
=
\frac{1-q_i^{\ell_i}}{-\ln q_i}.
\]
Therefore, the expected reachable length along the candidate path is
\[
\hat{L}
=
\sum_{i=1}^{N} R_i d_i
=
\sum_{i=1}^{N}
\left(
\prod_{j=1}^{i-1} q_j^{\ell_j}
\right)
\frac{1-q_i^{\ell_i}}{-\ln q_i}.
\]
We then convert \(\hat{L}\) into a point \(\bar{p}\) on the path.
Let
\[
C_k=\sum_{i=1}^{k}\ell_i, \qquad C_0=0
\]
denote the cumulative path length.
We find the segment index \(k\) satisfying
\[
C_{k-1}\le \hat{L}\le C_k,
\]
and compute the interpolation ratio
\[
\alpha=\frac{\hat{L}-C_{k-1}}{\ell_k}.
\]
The expected reach point is obtained by linear interpolation:
\[
\bar{p}=(1-\alpha)p_{k-1}+\alpha p_k.
\]

Finally, each candidate path is scored using its expected reach point.
Given a goal position \(g_t\in\mathbb{R}^2\), we define the goal advance and displacement scores as
\begin{align*}
s_{\mathrm{goal}}
&=
\mathrm{clip}\left(
\frac{\|p_0-g_t\|_2-\|\bar{p}-g_t\|_2}{\rho_{\mathrm{goal}}},
-\infty,
1
\right), \\
s_{\mathrm{disp}}
&=
\mathrm{clip}\left(
\frac{\|\bar{p}-p_0\|_2}{\rho_{\mathrm{disp}}},
0,
1
\right).
\end{align*}
Here, \(s_{\mathrm{goal}}\) measures the reduction in distance to the goal,
while \(s_{\mathrm{disp}}\) measures how far $\bar{p}$ lies from the current position.

To reduce temporal oscillation, we additionally introduce a temporal consistency score with respect to the previously selected path.
Let \(p_{t-1,i}\) denote the \(i\)-th waypoint of the previously selected path, expressed in the current robot frame.
We define \(\Delta\phi_i\) as the difference in bearing angle between \(p_i\) and \(p_{t-1,i}\).
The temporal consistency score is
\[
s_{\mathrm{temp}}
=
\mathrm{clip}
\left(
1-
\frac{1}{N}
\sum_{i=1}^{N}
\frac{|\Delta\phi_i|}{\phi_{\mathrm{temp}}},
0,
1
\right).
\]

The final score for each candidate path is
\[
s
=
\lambda_{\mathrm{goal}}\,s_{\mathrm{goal}}
+
\lambda_{\mathrm{disp}}\,s_{\mathrm{disp}}
+
\lambda_{\mathrm{temp}}\,s_{\mathrm{temp}}.
\]
We select the candidate path with the highest score as the final path.
In our experiments, we set
\(\rho_{\mathrm{goal}}=\rho_{\mathrm{disp}}=5\),
\(\phi_{\mathrm{temp}}=\pi/4\),
\(\lambda_{\mathrm{goal}}=0.6\),
\(\lambda_{\mathrm{disp}}=0.4\), and
\(\lambda_{\mathrm{temp}}=0.2\).

\begin{figure*}[t]
    \centering
    \includegraphics[width=\textwidth]{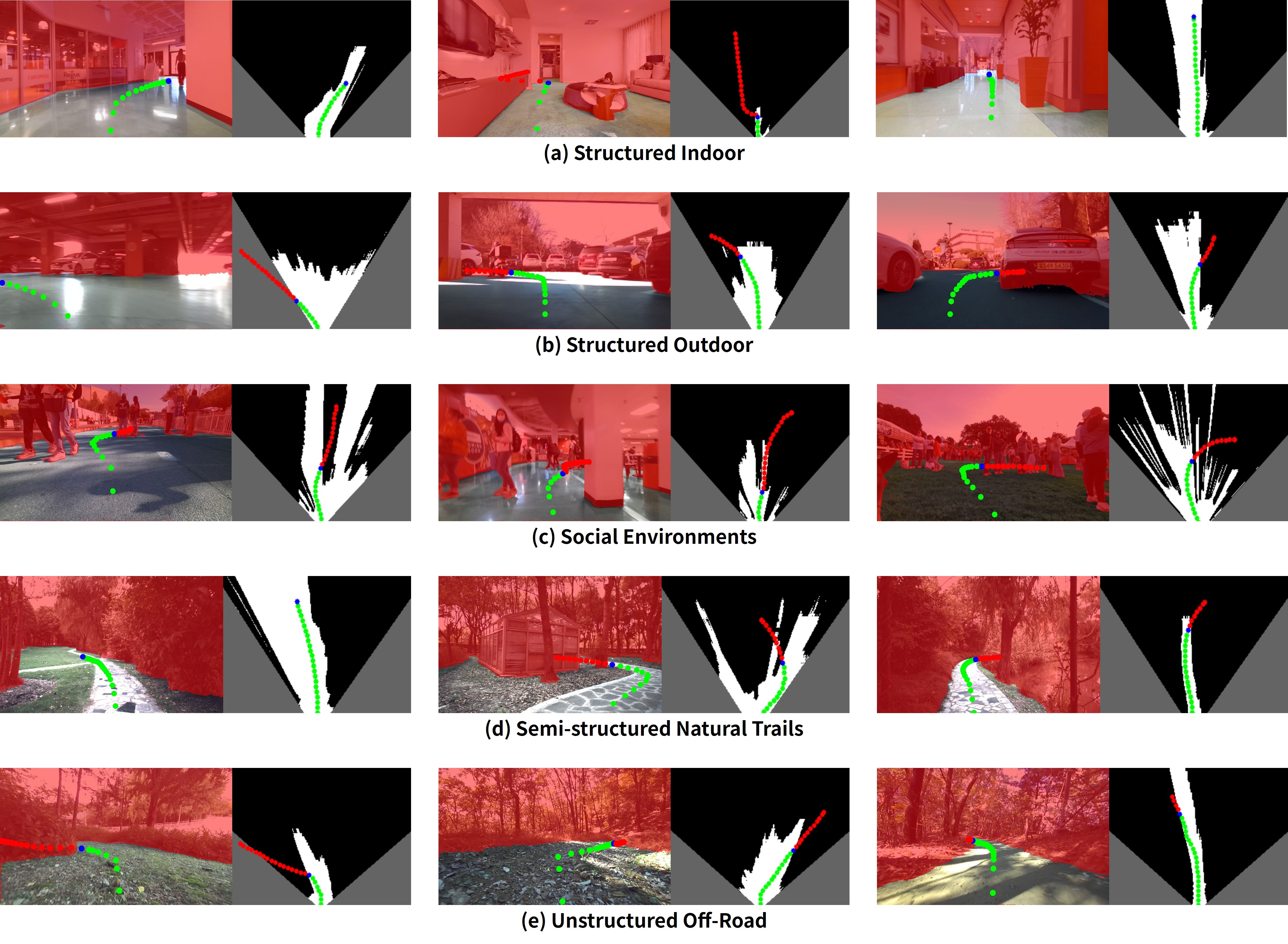}
    \caption{Pseudo-label generation across diverse environments. Each example shows the predicted traversability mask, the corresponding BEV collision map, and the expert trajectory. The blue point denotes the Scope-of-Reach target.}
    \label{supp:data_preprocess}
\end{figure*}

\subsection{Pseudo-Label Generation for Training}
For each training image, a frozen offline traversability
estimator~\citep{hwang2026general} predicts a per-pixel
traversability mask. This estimator is used only during dataset
preprocessing and is not required at inference time.

We construct a robot-centric collision map
\(M_{\mathrm{coll}}\) on an assumed flat ground plane. The map extends
\(32\,\mathrm{m}\) to either side of the robot and
\(32\,\mathrm{m}\) in the forward direction, with a resolution of
\(0.1\,\mathrm{m}\) per cell. Using the known camera geometry, we transform the center of each BEV cell into the camera
frame and project it onto the image plane. We then sample the
predicted traversability mask at the projected pixel and mark the
corresponding BEV cell as occupied if the pixel is classified as
non-traversable. BEV cells whose projected locations fall within the near-field blind region are excluded when constructing the SoR target.

To determine the SoR target, we first sample the expert trajectory
at \(0.33\,\mathrm{m}\) intervals and transform the sampled poses into
the local BEV frame. We then evaluate
whether each pose lies in a collision region of
\(M_{\mathrm{coll}}\). Checking only the single BEV cell corresponding to the robot center
can produce unreliable collision labels due to noise in the
estimated traversability mask. We therefore evaluate all BEV cells covered by the robot footprint at each pose. The footprint size is
set according to the physical dimensions
of the robot used for data collection. A pose is labeled as colliding
if at least \(30\%\) of the cells within its footprint are occupied. Starting
from the current pose, we follow the sampled expert trajectory until a
sampled pose either enters a collision region or exits the camera field
of view. The last valid pose before either event is
selected as the Scope-of-Reach target.

Figure~\ref{supp:data_preprocess} visualizes the predicted
traversability masks, their BEV projections, and the resulting SoR
targets across diverse environments. For visualization purposes only,
we crop the BEV map to \(12\,\mathrm{m}\) on each side of the robot
and \(18\,\mathrm{m}\) in the forward direction, and resample the
expert trajectories at \(0.7\,\mathrm{m}\) intervals.

\begin{table*}[t]
\centering
\footnotesize
\setlength{\tabcolsep}{5pt}
\renewcommand{\arraystretch}{1.3}

\begin{tabular*}{\textwidth}{
    @{\extracolsep{\fill}}
    l l l l
    @{}
}
\toprule
\textbf{Module}
& \textbf{Input}
& \textbf{Architecture}
& \textbf{Output} \\
\midrule

DINOv3 backbone
& $\tilde I_t$: $3 \times 256 \times 336$
& ViT-S+ with LoRA, layers $\{3,6,9,12\}$
& $\{f_I^{(\ell)}\}$:
  $4 \times (384\times16 \times 21)$ \\

\midrule 

Feature aggregation
& $\{f_I^{(\ell)}\}$:
  $4 \times (384 \times 16 \times 21)$
& \makecell[l]{$4 \times$ Conv$_{1\times1}$ $(384 \rightarrow 96)$ \\
  Concatenation ($96 \times 4 \rightarrow 384$) \\
  Conv$_{1\times1}$ ($384 \rightarrow 256$)}
& $f_I$: $256 \times 16 \times 21$ \\

\midrule

Goal encoder
& $\tilde g_t$: $2$
& MLP: $2 \rightarrow 256$
& $f_G$: $256$ \\

\midrule

Cross-attention
& \makecell[l]{
    $f_G$: $256$ \\
    $f_I$: $256 \times 16 \times 21$
  }
& \makecell[l]{
    Multi-head attention \\
    Query: $f_G$, Key \& Value: $f_I$
  }
& $f_A$: $256$ \\

\midrule

Feature concatenation
& \makecell[l]{
    $f_I$: $256 \times 16 \times 21$ \\
    $f_G$: $256$ \\
    $f_A$: $256$
  }
& \makecell[l]{
    Global average pooling of $f_I$ to obtain $f_I^{pool}$\\
    Concatenate $[f_I^{pool}, f_G, f_A]$
  }
  
& $f_{\mathrm{cond}}$: $768$ \\

\midrule

Scope-of-Reach head
& $f_{\mathrm{cond}}$: $768$
& \makecell[l]{
    Shared MLP: $768 \rightarrow 256 \rightarrow 256$ \\
    Mean MLP: $256 \rightarrow 2$ \\
    Std MLP: $256 \rightarrow 2$
  }
  
& \makecell[l]{
    $\tilde{\mu}$: 2 \\
    $\tilde{\sigma}$: 2} \\

\midrule

Diffusion policy
& \makecell[l]{
    $\tilde{\tau}_k$: $N \times 2$ \\
    $k$: $1$ \\
    $f_{\mathrm{cond}}$: $768$
  }
&
\makecell[l]{
Conditional 1D U-Net \\
($2 \rightarrow 128 \rightarrow 256 \rightarrow 512
\rightarrow 256 \rightarrow 128 \rightarrow 2$) \\
Conditioned on $k$ and $f_{\mathrm{cond}}$ at each block \\
Total denoising steps: 10}
& $\epsilon_\theta(\tilde{\tau}_k,k,f_{\text{cond}})$: $N \times 2$ \\

\midrule
Traversability head
& \makecell[l]{
    $\tilde{\tau}_0$: $N \times 2$ \\
    $f_I$: $256 \times 16 \times 21$
  }
&
\makecell[l]{
Sample $f_I$ at the waypoints in $\tilde{\tau}_0$: $N \times 256$ \\
Concatenate with $\tilde{\tau}_0$, followed by MLP and Conv1D
}
& $\hat{q}$: $N$ \\

\bottomrule
\end{tabular*}
\caption{Network architecture details.}
\label{tab:network_details}
\end{table*}

\section{Network and Training Details}
\subsection{Network Architecture}
The detailed network architecture is summarized in Table~\ref{tab:network_details}. CanonNav extracts image features using a DINOv3~\citep{simeoni2025dinov3} ViT-S+ backbone adapted with LoRA~\citep{hu2022lora}. We aggregate intermediate features
from the 3rd, 6th, 9th, and 12th transformer layers to capture both local details and global scene context. The encoded goal feature serves as the query in a cross-attention module, while the image features serve as the keys and values. The resulting conditioning feature is then provided to the subsequent prediction heads.

Following prior work~\citep{sridhar2024nomad}, we implement the
diffusion policy using a Conditional 1D U-Net, with the fused feature
provided as a conditioning signal. During training, we randomly sample a diffusion timestep
$k$ and Gaussian noise, and corrupt the expert trajectory $\tau_0$
according to the corresponding noise level to obtain $\tau_k$. The
policy then predicts the added noise given $\tau_k$. Using the
predicted noise, we recover a clean trajectory estimate
$\hat{\tau}_0$ following Eq.~(6) of the main paper and pass it to
the traversability head. For real-time navigation, the traversability head
instead receives the final trajectory produced by the full
10-step denoising process.

\begin{table}[t]
\centering
\footnotesize
\setlength{\tabcolsep}{4pt}
\renewcommand{\arraystretch}{1.15}

\begin{tabularx}{\columnwidth}{
    @{}
    >{\centering\arraybackslash}X
    >{\centering\arraybackslash}p{0.20\columnwidth}
    >{\centering\arraybackslash}p{0.22\columnwidth}
    @{}
}
\toprule
\textbf{Name}
& \textbf{Symbol}
& \textbf{Value} \\
\midrule
Canonical intrinsics
& \makecell{
    $\tilde{f}_x, \tilde{f}_y, \tilde{c}_x $ \\
    $\tilde{c}_y$
    }
& \makecell{
    $168$ \\
    $128$}
\\

\midrule

Number of waypoints 
& $N$
& $16$ \\
\midrule

Number of trajectory candidates
& $B$
& $30$ \\
\midrule

Robot radius
& $d_{\mathrm{robot}}$
& $0.25$ \\
\midrule
Safety margin
& $d_{\mathrm{safe}}$
& $0.75$ \\
\midrule

Loss function sharpness
& \makecell{
    $\eta$ \\
    $\gamma$ \\
    $\rho$
    }
& \makecell{
    $0.1$ \\
    $2$ \\
    $0.5$
    }
\\

\midrule

Loss weights
& \makecell{
    $\lambda_{\mathrm{diff}}$ \\
    $\lambda_{\mathrm{coll}}$ \\
    $\lambda_{\mathrm{trav}}$ \\
    $\lambda_{\mathrm{pred}}$ \\
    $\lambda_{\mathrm{neg}}$ \\
    $\lambda_{\mathrm{cons}}$ \\
    $\lambda_{\mathrm{smooth}}$
    } 

& \makecell{
    $25.0$ \\
    $0.3$ \\
    $5.0$ \\
    $2.0$ \\
    $8.0$ \\
    $0.25$ \\
    $0.05$} \\
\bottomrule
\end{tabularx}
\caption{Model hyperparameters.}
\label{tab:hyperparameters}
\end{table}

\subsection{Training Details}
The hyperparameters used in our experiments are summarized in
Table~\ref{tab:hyperparameters}. During training, we apply random yaw augmentation and horizontal
flipping. For yaw augmentation, we sample an angle from
$[-15^\circ,15^\circ]$ and replace
$R_{\mathrm{pitch}}^{-1}$ in Eq.~(3) of the main paper with
$R_{\mathrm{yaw}}R_{\mathrm{pitch}}^{-1}$. For each augmentation, the corresponding transformation is applied to both
the image and the expert trajectory to preserve geometric consistency.

We use LoRA adapters with a rank of 3 and optimize their parameters with
a learning rate of $1\times10^{-4}$. All remaining trainable
parameters are optimized with a learning rate of $5\times10^{-3}$.
We use the AdamW optimizer and train the model for 70 epochs with a total batch
size of 256. After each optimization step, we update an
exponential moving average (EMA) copy of the model using a warm-up schedule that increases the EMA decay up to $0.9999$. We use the EMA model for real-time navigation. At each inference step, the policy generates 30 candidate trajectories,
each consisting of 16 waypoints, and selects the final path using the
path-selection procedure described in Sec.~\ref{supp_section:path_selection}.

\subsection{Embodiment-Specific Planning Parameters}
The robot radius \(d_{\mathrm{robot}}\) and safety margin
\(d_{\mathrm{safe}}\) in Table~\ref{tab:hyperparameters} are embodiment-specific parameters defined for
the target platform. The robot radius is used in the traversability
loss \(\mathcal{L}_{\mathrm{trav}}\) to determine whether a waypoint
overlaps an obstacle, while the safety margin introduces additional
clearance through the collision loss \(\mathcal{L}_{\mathrm{coll}}\),
encouraging generated trajectories to remain farther from unsafe
regions. Together, these parameters allow the trained policy to
reflect both the physical size of the deployment platform and the
desired obstacle clearance. These embodiment-specific parameters are fixed during training, requiring the policy to be retrained with the corresponding settings for each new deployment platform. Learning a unified cross-embodiment policy that adapts its navigation behavior according to embodiment-specific parameters remains an important direction for future work.

\section{Additional Qualitative Results}
\subsection{Simulation Environments and Results}
\begin{figure*}[!p]
    \centering
    \includegraphics[width=0.9\textwidth]{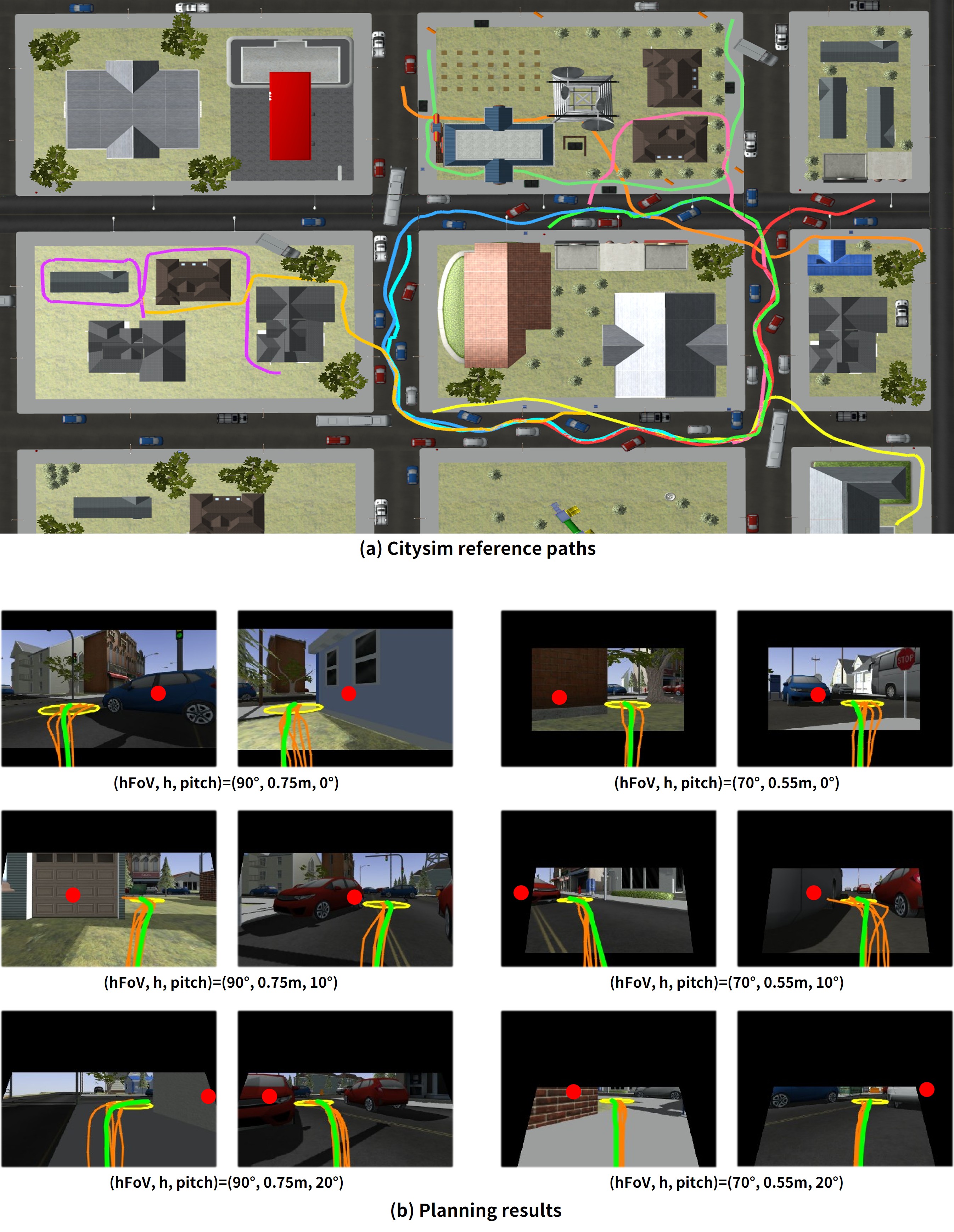}
    \caption{CitySim environment, reference paths, and planning results. (a) Full map and reference paths used for evaluation, with each path shown in a different color. (b) Representative planning results across different platform settings.}
    \label{supp:citysim}
\end{figure*}

\begin{figure*}[p]
    \centering
    \includegraphics[width=0.9\textwidth]{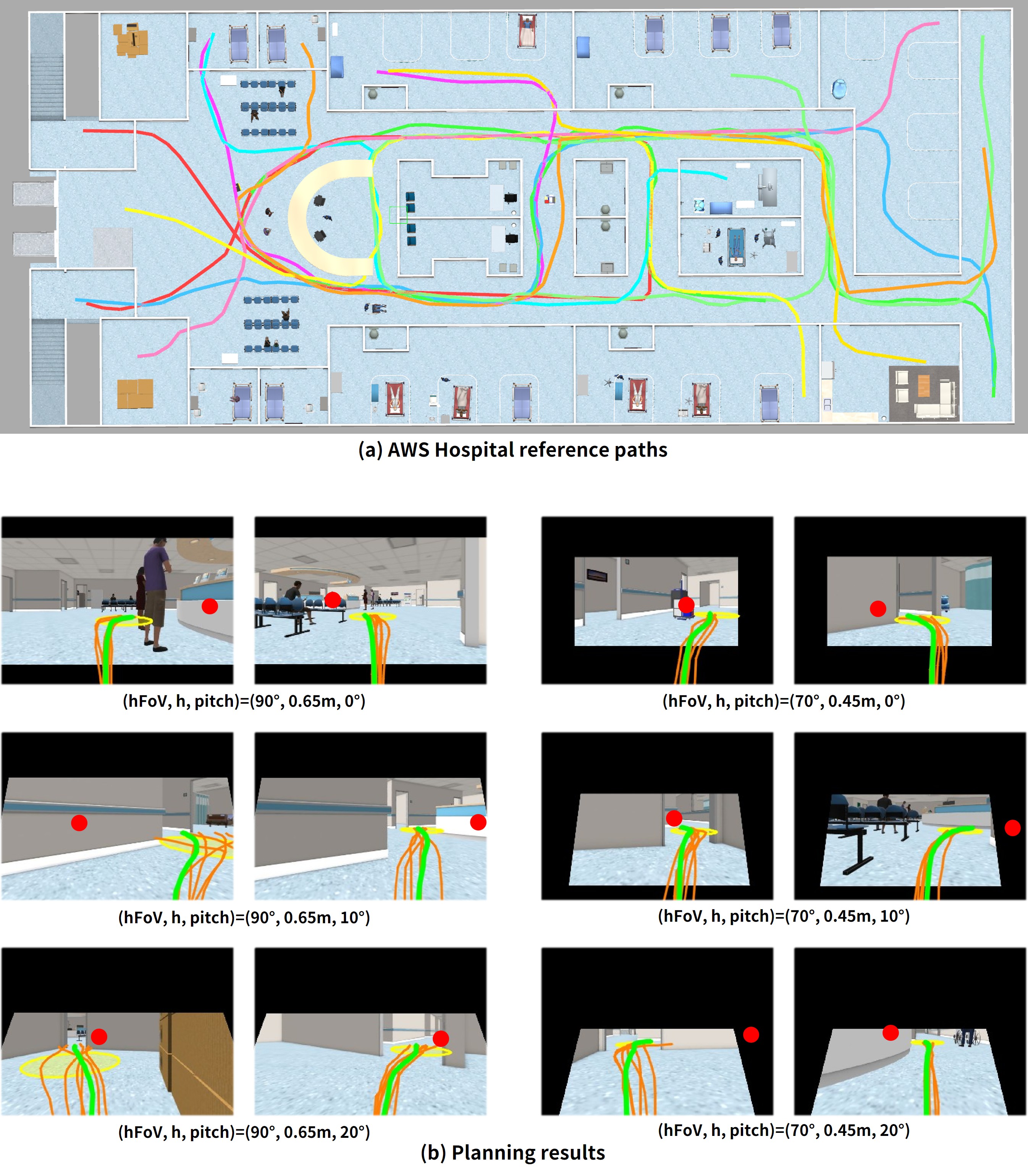}
    \caption{AWS Hospital environment, reference paths, and planning results. (a) Full map and reference paths used for evaluation, with each path shown in a different color. (b) Representative planning results across different platform settings.}
    \label{supp:aws_hospital}
\end{figure*}

Figs.~\ref{supp:citysim} and~\ref{supp:aws_hospital} illustrate the simulation environments used in the main paper. For each figure, the top panel shows the full map and the reference paths used for evaluation, with each colored curve denoting a distinct reference path. The bottom panel presents representative planning results across different platform settings, with variations in the camera's horizontal field of view, mounting height, and pitch angle. To increase the difficulty of the CitySim experiments, we placed additional objects throughout the environment to create cluttered areas and narrow traversable passages. The reference paths were designed to pass through these regions, requiring the policy to navigate around nearby obstacles and make precise local planning decisions.

\subsection{Real-world Environments and Results}
\begin{figure*}[p]
    \centering
    \includegraphics[width=0.96\textwidth]{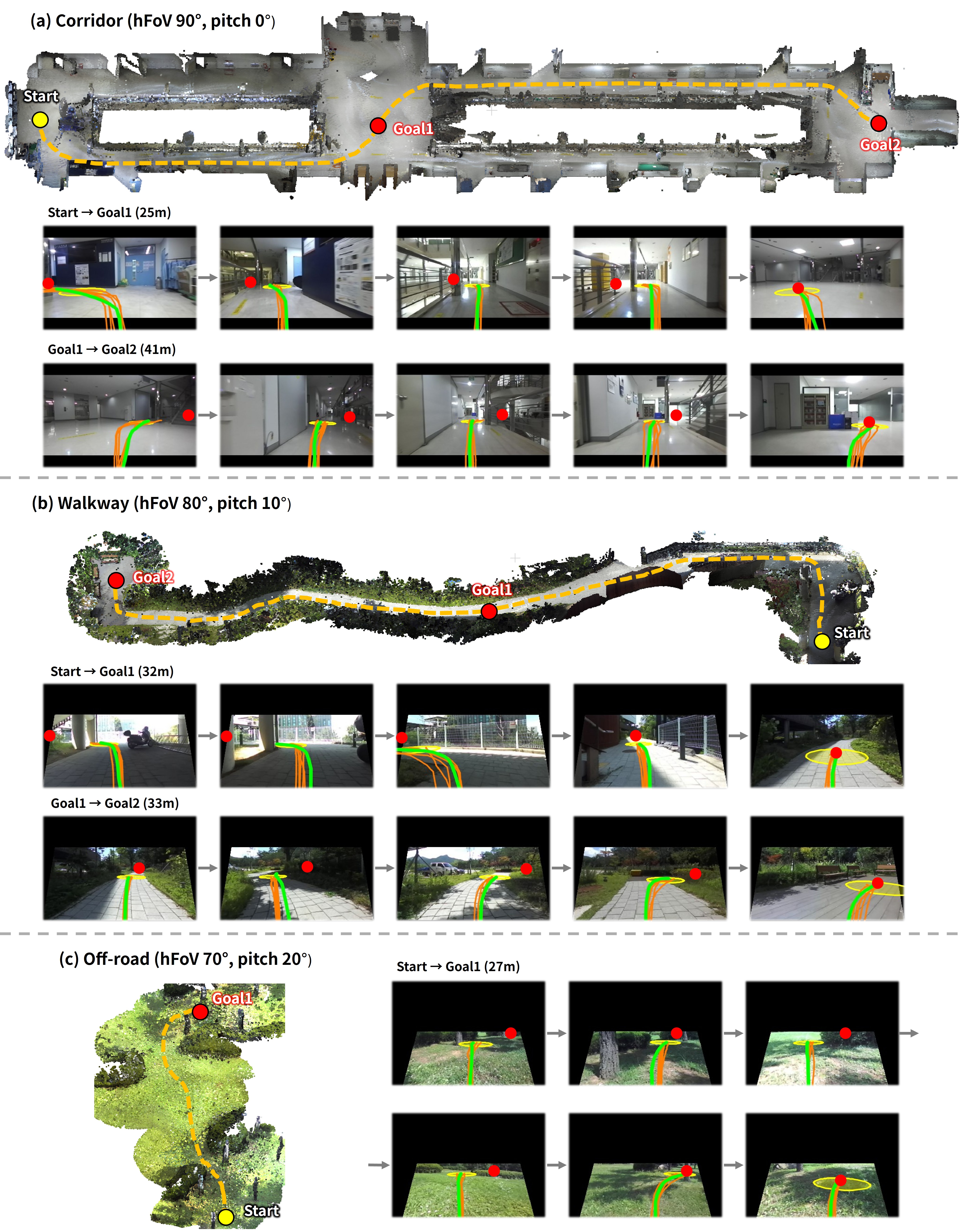}
    \caption{Full qualitative results from the real-world experiments. For each environment, the reconstructed top-down map shows the complete trajectory, start position, and subgoals. The planning results are grouped by subgoal segment, with gray arrows indicating the temporal order within each segment.}
    \label{supp:real_world}
\end{figure*}

Fig.~\ref{supp:real_world} presents the full set of qualitative results from the real-world experiments reported in the main paper. For each environment, the reconstructed top-down map shows the full trajectory executed by the robot as a dashed yellow curve, together with the start position and subgoals. The accompanying images present representative planning results along each subgoal segment, with the gray arrows indicating their temporal order. These results show how the policy continuously updates its local trajectory
while navigating in diverse indoor, outdoor, and unstructured environments
under different camera configurations.

\subsection{Failure Cases of CanonNav}
\begin{figure*}[t]
    \centering
    \includegraphics[width=0.7\textwidth]{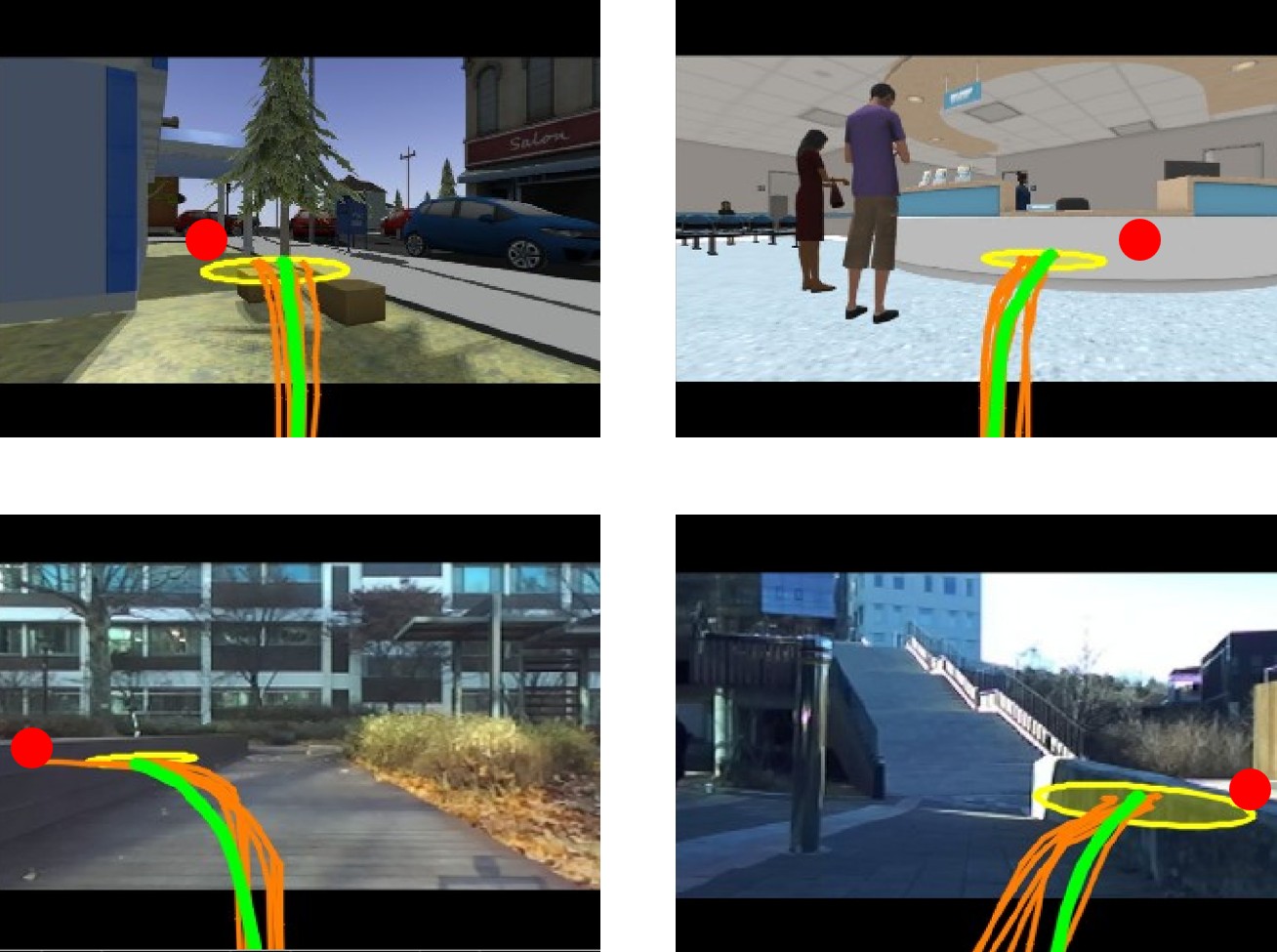}
    \caption{Representative failure cases of CanonNav. Most failures arise from incorrect Scope-of-Reach predictions. These failures typically occur when obstacles are poorly represented in the training distribution (e.g., the small box in the upper-left example) or are visually difficult to distinguish from their surroundings.}
    \label{supp:failure}
    \vspace{10pt}
\end{figure*}

Fig.~\ref{supp:failure} shows representative failure cases of CanonNav.
Most of these failures arise from incorrect Scope-of-Reach (SoR) predictions.
In these cases, the predicted SoR overlaps with a non-traversable region, causing the policy to generate trajectories toward that area.
These failures mainly occur when obstacles are poorly represented in the
training distribution (e.g., the small box in CitySim) or are difficult
to distinguish because of limited texture cues or visual similarity to
the surrounding scene.
Despite these failure modes, our quantitative and qualitative results
demonstrate CanonNav's robust generalization across diverse camera
configurations and environments.

